\documentclass[twoside,leqno,twocolumn]{article}

\usepackage[letterpaper]{geometry}

\usepackage{ltexpprt}
\usepackage{hyperref}

\usepackage[square,numbers]{natbib}
\usepackage{amsmath}
\usepackage{soul}
\usepackage{xcolor}
\usepackage[export]{adjustbox}
\usepackage{hyperref}
\usepackage{listings}
\usepackage{bm}
\usepackage{array,multirow,graphicx}
\usepackage{diagbox}
\usepackage[
  separate-uncertainty = true,
  multi-part-units = repeat
]{siunitx}
\usepackage{enumitem}
\usepackage{makecell}
\usepackage{lipsum}
\let\OLDthebibliography\thebibliography
\renewcommand\thebibliography[1]{
  \OLDthebibliography{#1}
  \setlength{\parskip}{0pt}
  \setlength{\itemsep}{0pt plus 0.3ex}
}

\begin{document}

\newcommand\relatedversion{}

\title{\Large Mini-Batch Learning Strategies for modeling long term temporal dependencies: A study in environmental applications\relatedversion}
\author{Shaoming Xu \thanks{University of Minnesota.  \{xu000114, khand035, lixx5000, lichengl, willa099, ghosh128, lind0436, stei0062, nieber,kumar001\}@umn.edu}
\and Ankush Khandelwal \footnotemark[1]
\and Xiang Li \footnotemark[1]
\and Xiaowei Jia\thanks{University of Pittsburgh. xiaowei@pitt.edu}
\and Licheng Liu \footnotemark[1]
\and Jared Willard \footnotemark[1]
\and Rahul Ghosh \footnotemark[1]
\and Kelly Cutler \footnotemark[1]
\and Michael Steinbach \footnotemark[1]
\and Christopher Duffy\thanks{Pennsylvania State University. cxd11@psu.edu}
\and John Nieber\footnotemark[1]
\and Vipin Kumar \footnotemark[1]
}

\date{}
\maketitle

% Copyright Statement
% When submitting your final paper to a SIAM proceedings, it is requested that you include
% the appropriate copyright in the footer of the paper.  The copyright added should be
% consistent with the copyright selected on the copyright form submitted with the paper.
% Please note that "20XX" should be changed to the year of the meeting.

% Default Copyright Statement
\fancyfoot[R]{\scriptsize{Copyright \textcopyright\ 2023 by SIAM\\
Unauthorized reproduction of this article is prohibited}}

% Depending on which copyright you agree to when you sign the copyright form, the copyright
% can be changed to one of the following after commenting out the default copyright statement
% above.

%\fancyfoot[R]{\scriptsize{Copyright \textcopyright\ 20XX\\
%Copyright for this paper is retained by authors}}

%\fancyfoot[R]{\scriptsize{Copyright \textcopyright\ 20XX\\
%Copyright retained by principal author's organization}}

%\pagenumbering{arabic}
%\setcounter{page}{1}%Leave this line commented out.
\begin{abstract} \label{abstract}
\small
In many environmental applications, recurrent neural networks (RNNs) are often used to model physical variables with long temporal dependencies. However, due to mini-batch training, temporal relationships between training segments within the batch (intra-batch) as well as between batches (inter-batch) are not considered, which can lead to limited performance. Stateful RNNs aim to address this issue by passing hidden states between batches. Since Stateful RNNs ignore intra-batch temporal dependency, there exists a trade-off between training stability and capturing temporal dependency.  In this paper, we provide a quantitative comparison of different Stateful RNN modeling strategies, and propose two strategies to enforce both intra- and inter-batch temporal dependency. First, we extend Stateful RNNs by defining a batch as a temporally ordered set of training segments, which enables intra-batch sharing of temporal information. While this approach significantly improves the performance, it leads to much larger training times due to highly sequential training. To address this issue, we further propose a new strategy which augments a training segment with an initial value of the target variable from the timestep right before the starting of the training segment. In other words, we provide an initial value of the target variable as additional input so that the network can focus on learning changes relative to that initial value. By using this strategy, samples can be passed in any order (mini-batch training) which significantly reduces the training time while maintaining the performance. In demonstrating the utility of our approach in hydrological modeling, we observe that the most significant gains in predictive accuracy occur when these methods are applied to state variables whose values change more slowly, such as soil water and snowpack, rather than continuously moving flux variables such as streamflow.
\end{abstract}

\section{Introduction}
RNNs have been widely used in many time series applications such as healthcare \cite{shickel2017deep}, finance \cite{sezer2020financial}, hydrology \cite{sit2020comprehensive}, and weather \cite{mauro2022harnessing}. The key innovation behind the success of RNNs is the joint use of Back Propagation Through Time (BPTT) \cite{werbos1990backpropagation} and mini-batch training (MB) \cite{hinton2012neural}. Specifically, by breaking very long timeseries datasets into small training segments (samples), we can efficiently train RNNs on large datasets. RNNs can also be enhanced with architectures such as LSTM \cite{hochreiter1997long}  and GRU \cite{cho2014learning} to address the issue of vanishing gradients \cite{pascanu2013difficulty}. While BPTT and MB enable the practical use of RNNs, they diminish the use of long temporal dependencies during model training. BPTT on smaller segments and the independent and identically distributed assumption on samples in traditional mini-batch training ignore the temporal relationship between training segments and reduce the access to longer history to train models. This leads to reduced performance in certain applications such as soilwater, the amount of water in the soil,  whose temporal dependency structure can span multiple years. 

Stateful RNNs \cite{gulli2017deep} were proposed to address this problem partially. Specifically, Stateful RNNs pass hidden states from one batch to another. While this introduces temporal dependency across batches (inter-batch), Stateful RNNs still consider each segment within the batch to be independent. Consequently, researchers have to reduce the batch size and set more batches for long-term temporal information, leading to training instability. For example, by setting the batch size as one like stochastic gradient descent, Stateful RNNs can use complete temporal information at the expense of highly unstable training and lower convergence rate\cite{bottou2007tradeoffs}. In this paper, we propose two new approaches to prepare mini-batches that incorporate both intra- and inter-batch dependencies for improving the performance of RNNs while leveraging the computational benefits of mini-batch training and BPTT. 

First, we extend Stateful RNNs to enable the intra-batch sharing of temporal information by defining a batch as a temporally ordered set of training segments. During the forward pass for a batch, we pass the detached hidden states between segments, and we backpropagate the average loss of the batch to update the models' weights. We show that this extension can achieve significant performance improvements. 

Despite the achieved performance improvement, this strategy has two key issues. First,  if the last few batches are anomalous compared to the rest of the dataset, the models can overfit the batches and thus get optimized in the wrong direction. Second, the training time of this method is much longer than traditional MB. This is because the strategy cannot fully leverage the matrix acceleration of modern GPUs due to its sequential nature. 

To address these drawbacks, we propose a new strategy to prepare mini-batches, and this strategy is inspired by the modeling of bio-physical systems. In scientific simulations, the initial state of the physical system is provided as input and the simulation changes these states based on the observed inputs. In other words, the simulation does not infer the state directly but simulates the changes to the state from a given starting point. We propose to use the same strategy by providing an initial value as an additional input for each training segment. For an RNN training segment, the initial value will be the value of the target variable from the timestep before the starting timestep of the segment. From a machine learning perspective, this can be seen as a variation of the teacher forcing strategy \cite{williams1989learning}. In traditional teacher forcing, each timestep uses the ground truth from the previous timestep. While this has shown improvement in certain applications, it introduces a serious issue in modeling physical systems. Passing the previous timestep's ground truth at every timestep, the model only needs to learn the change in the state for that step and hence cannot learn to accumulate these changes correctly over time. While this shows good improvement during training, it leads to significant error propagation during inference. Hence, we propose to use the teacher forcing concept only for the first timestep of each sample to allow the model to capture state changes over time. 

While the concept of passing hidden states can be applied to any application of RNNs, the idea of using an initial target value as input can only be used where the target variable represents cumulative effects on inputs on the target variable. As a counterexample, in language translation applications, passing hidden states can communicate more about the previous sentence than just passing the last word as input for the next RNN segment. Hence, using target values as inputs is better suited for quantitative applications such as hydrology, weather, and finance. 

We show the efficacy of the proposed strategies on three physical variables related to Earth's hydrological cycle, including soil moisture (soilwater), snowpack, and streamflow. Among these three variables, soilwater and snowpack are state variables whose temporal dependency spans multiple years and seasons respectively. Streamflow is a flux variable that can largely be explained by weather inputs for a given day. Hence, these three variables provide a good framework for analyzing different training strategies. 

The contributions of this paper are summarized as follows:
\begin{itemize}
    \itemsep0em 
    \item We propose two new training algorithms to prepare mini-batches for improving the performance of RNNs on tasks with long temporal dependencies.
    \item We pair the training algorithms with three inference algorithms to get five learning strategies and assess the impact of temporal dependency choice on their performance.
    \item We demonstrate the efficacy of our proposed strategies on the societally important and challenging problem of modeling hydrological variables using RNNs.
    \item We share the data and the proposed algorithms in the GitHub repository\footnote{\url{https://github.com/XuShaoming/MiniBatch_Learning_Strategies}}.
\end{itemize}

\section{Problem Formulation}
We design learning algorithms that help sequential models (e.g., RNNs) to learn a function that maps the $i$th input segment $X^i = (X^i_1,\dots,X^i_T)$ to the $i$th target segment $Y^i=(Y^i_1, \dots, Y^i_T)$ of the same length. Table \ref{tab:notation} summarizes the notation used in this paper.

\begin{table}[h]
    \caption{notation used in this paper.}
    \label{tab:notation}
    \resizebox{\columnwidth}{!}{% Resize table to fit within
        \begin{tabular}{ |c|c|}
            \hline
            $X^i$ & The $i$-th input segment of length $T$, $X^i = (X^i_1,\dots,X^i_T)$\\
            \hline
             $Y^i$ & The $i$-th target segment of length $T$, $Y^i = (Y^i_1,\dots,Y^i_T)$\\
            \hline
            $\hat{Y}^i$ & The $i$-th predicted target segment of length $T$, $\hat{Y}^i = (\hat{Y}^i_1,\dots,\hat{Y}^i_T)$\\
            \hline
            $Y^{i}_0$ & The initial target of $i$th segment.\\
            \hline
            $\hat{Y}^{i}_0$ & The predicted initial target of $i$th segment.\\
            \hline
            $T$ & Length of a segment.\\
            \hline
            $H^i_t$ & The hidden states of $i$th segment at $t$th time step.\\
            \hline
            $hs$ & hidden states size\\
            \hline
            $bs$ & The number of segments in a batch\\
            \hline
            $\vec{a}_{[n]}$ & A single numeric vector of length $n$\\
            \hline
        \end{tabular}
    }
\end{table}
\section{Baseline Mini-Batch algorithms}
\begin{figure*}[t]
\centering
\includegraphics[width=\textwidth]{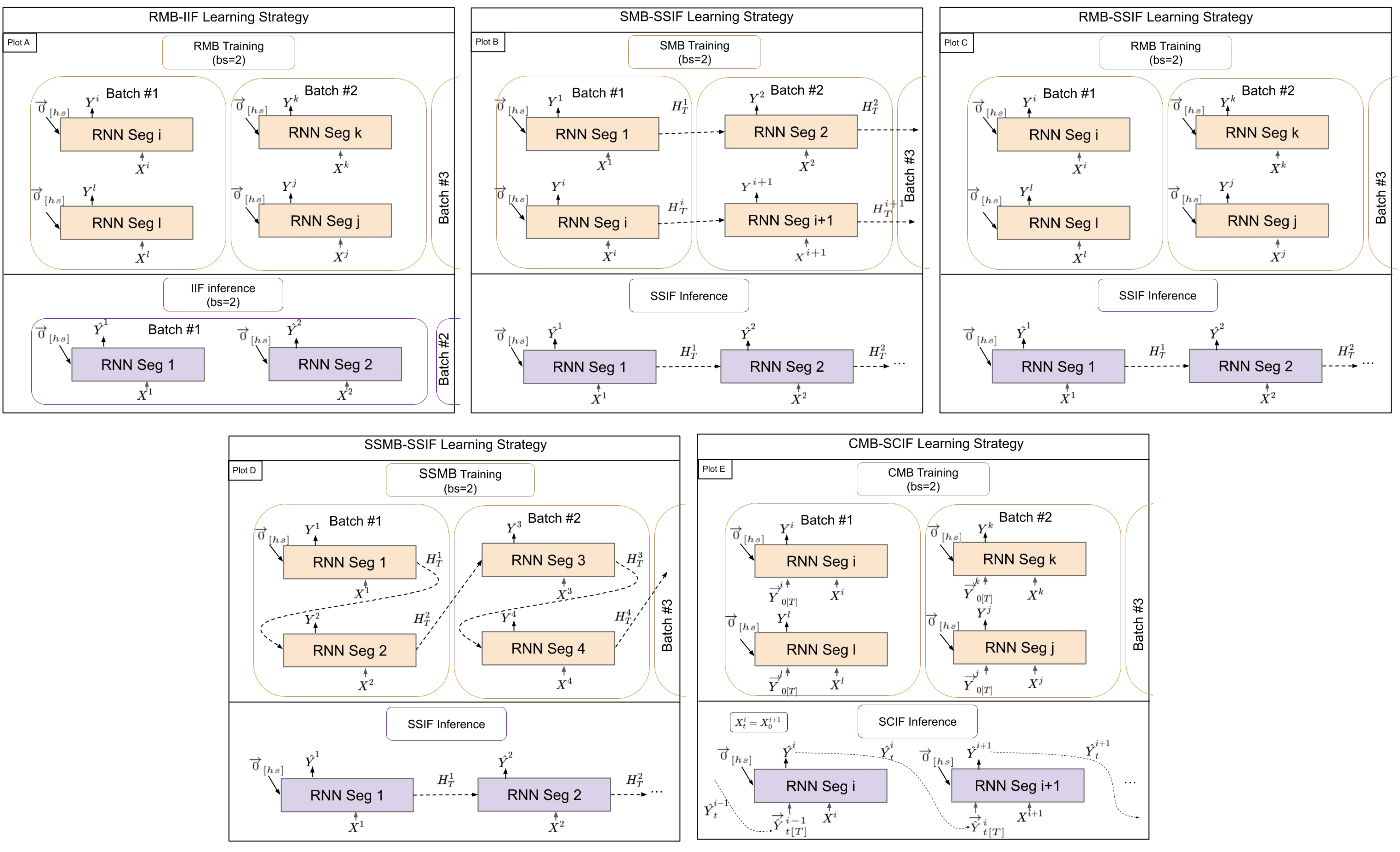} % Reduce the figure size so that it is slightly narrower than the column.
\caption{This figure shows the learning strategies. CMB-SCIF learning can overlap the segments where the $t$th step of the $i$th segment matches the $0$th step of the $i+1$th segment. We set $bs=2$ to make figure easy to read. Losses can not backpropagate over dash lines.}
\label{fig:learning_methods}
\end{figure*}

\begin{table}[t]
    \caption{The learning strategies as the combination of training (TR) and inference (IF) algorithms.}
    \label{tab:opt_pred_combination}
    \resizebox{\columnwidth}{!}{% Resize table to fit within \linewidth horizontally
% \begin{tabular}{ |p{3cm}|p{3cm}|p{3cm}|p{3cm}|p{3cm}|}
    \begin{tabular}{|l|*{3}{c|}}\hline
    % \backslashbox{training}{inference}
    \backslashbox{TR}{IF}
    &IIF  & SSIF  & SCIF\\
    \hline
    RMB &RMB-IIF  &RMB-SSIF&\\\hline
    SMB &&SMB-SSIF&\\\hline
    SSMB &&SSMB-SSIF&\\\hline
    CMB &&& CMB-SCIF\\\hline
    \end{tabular}
    }
\end{table}

\subsection{Random Mini-Batches (RMB).}
In RMB training algorithm (figure \ref{fig:learning_methods}, plot A), the training segments are randomly sampled without replacement and assigned to each batch; the model is trained on each batch until all batches are used; then, RMB starts a new epoch and trains the model over and over again until the model reaches the stopping criteria (e.g., the validation loss does not decrease over a certain number of epochs). 

RMB algorithm prevails in deep learning (DL) for several reasons. First, random sampling creates different training batches in each epoch, which helps the model escape from the local minima. Second, the mean loss in every batch can reduce noisy gradient signals and stabilize the training process. Third, the RMB algorithm can leverage the vectorized implementation provided by the CUDA Toolkit to accelerate computation. Lastly, RMB can control the batch size to train models on big data using limited computational resources (e.g., GPU and main memory). 

Although RMB has these advantages, it diminishes the use of temporal dependencies during training since it assumes each segment to be independent. Specifically, RMB initializes each RNN segment with zero hidden states, which ignores the valuable longer history and the fact that the consecutive segments in the time series are highly dependent. Thus, RMB can lead to poor performance for variables with long temporal dependencies.

\subsection{Stateful Mini-Batches (SMB).}
SMB training algorithm (figure \ref{fig:learning_methods}, plot B) passes RNN hidden states between batches, which makes the segments temporally dependent and respects the nature of time series data. In TensorFlow \cite{abadi2016tensorflow} and  Keras \cite{gulli2017deep}, we can activate SMB by setting the RNNs' parameter $stateful=True$; then, the algorithm will use the last hidden states of each RNN segment from a batch to initialize each segment at the same position in the next batch. 

Since SMB passes hidden states between batches, models are trained by only a limited amount of historical information. When the number of batches is small, each batch contains more "independent" segments, and fewer hidden states pass between batches. When the number of batches is large, more hidden states pass between batches. In the extreme case, when each batch only contains one segment, the number of batches equals the number of segments, and all historical information is used to train models. Though this one-at-a-time SMB training gains practical popularity \cite{yilmaz2019effect,elsworth2020time,katrompas2021enhancing}, it is can lead to unstable optimization and increased training time.

\section{Proposed Mini-Batch algorithms}
\subsection{Sequential Stateful Mini-Batches (SSMB).}
We design the SSMB algorithm (figure \ref{fig:learning_methods}, plot D) to train models using all historical information while maintaining more segments in each batch to stabilize the optimization. Strictly forcing the order of temporal segments to use during training, SSMB initializes the current RNN segment by the last hidden states of the previous RNN segment to enable the intra-batch sharing of temporal information. Moreover, we detach the last hidden states and only use their values to prevent passing errors between segments, which follows the Truncated Backpropagation Through Time algorithm \cite{williams1990efficient} to control the optimization complexity.

Though SSMB can stabilize the optimization by using the averaged gradients to update model weights, it requires a longer training time since it has to predict each segment individually to generate and pass hidden states sequentially. In addition, since both SMB and SSMB have to fix the order of segments to train models, they could overfit the models or optimize the models in the wrong direction if the last few batches are anomalous.

\subsection{Conditional Mini-Batches (CMB).}
In the CMB algorithm (figure \ref{fig:learning_methods}, plot E), the initial target value $Y_0^i$, observed one step before the first time step of $i$-th RNN segment, is copied at every time step and concatenated with $i$th input segment $X^i$ to get a new input segment $X^{'i}$; a new dataset is created. By providing the initial value, we made the training segments conditionally independent of each other because the initial value can be seen as a summary of the temporal effects of all previous timesteps. This conditional independence enables random shuffling of training segments and hence reduces training time significantly while maintaining good performance. 

CMB can be seen as a special variation of the teacher forcing strategy \cite{williams1989learning}. For variables that represent the state of a system, adjacent timesteps are often correlated. Traditional teacher forcing uses the observed target values from the previous timestep as the additional inputs of the current timestep, which makes the model unable to learn accumulated changes in the state over time during training and leads the model to significant error propagation during inference. CMB is designed to use only the initial target values to avoid these issues.

\section{Inference algorithms}
Temporal dependency during inference can also lead to changes in performance for any trained model. In this section, we describe three different inference algorithms and couple them with the training algorithms to form the learning strategies in table \ref{tab:opt_pred_combination}.

\subsection{Independent inference (IIF).}
In IIF (figure \ref{fig:learning_methods}, plot A), the trained model predicts each testing segment independently. IIF often serves as the default inference method of the RMB algorithm in practice, named RMB-IIF. Since this combination does not enforce temporal dependency during training as well as inference, it often leads to limited performance for long memory variables.

\subsection{Sequential stateful inference (SSIF).}
In SSIF (figure \ref{fig:learning_methods}, plots B, C, D), the trained model of the current segment is initialized by the hidden states from the previous segment to get all predictions in one pass. Since SSIF passes hidden states between every consecutive segment, the trained models can use all historical information to make predictions. The SSIF serves as the default inference method of SMB and SSMB training methods to get SMB-SSIF and SSMB-SSIF learning strategies respectively.

Moreover, we can apply SSIF on RMB-trained models to get the RMB-SSIF strategy. The RMB-SSIF works since RNN units share the same learned weights and can treat the previous segment's hidden states equally to the ones from the previous time step. Since RMB-SSIF not only leverages the long-term history but also avoids models' reconstruction of hidden states from scratch for each segment, RMB-SSIF can often provide superior prediction accuracy than RMB-IIF.  

\subsection{Sequential conditional inference (SCIF).}
The SCIF (figure \ref{fig:learning_methods}, plot E) is designed as the inference algorithm of CMB to get the CMB-SCIF learning strategy. In SCIF, we can provide a random initial target value (e.g., $Y_0^1 = 0$) as the state input to get the first prediction. Then the initial target value, as the state input of the current segment, is predicted from the previous segment until we get all predictions in temporal order.

Since the target values are already observed in the training and predicted in the inference, CMB-SCIF requires no additional observation and can be easily adapted to train existing models for diverse applications.

\section{Experiments} \label{sec:experiments}
\subsection{Dataset.} \label{sec:dataset}
\begin{figure}[tb]
    \centering
    \includegraphics[width =\columnwidth]{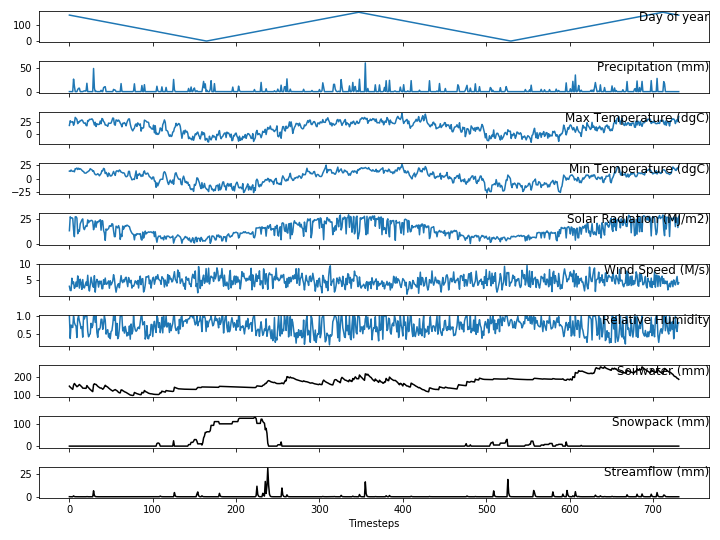}
    \caption{The blue time series represent 7 input features and the black time series represent 3 target features.}
    \label{fig:inputs_7feats_outputs_3feats}
\end{figure}

We demonstrate the effectiveness of different learning strategies on a hydrology dataset simulated by The Soil \& Water Assessment Tool (SWAT) \cite{arnold2012swat}. The dataset contains six input weather variables: precipitation, minimum daily temperature, maximum daily temperature, solar radiation, wind speed, and relative humidity. To include the seasonal information, we select the day of year (DOY), transformed by $183 - |doy - 183|$ to reflect the distance between days, as an additional input. To evaluate the performance of our proposed methods, we use three different yet important hydrological system output variables, namely Soilwater (SW), Snowpack (SNO), Streamflow (SF), with varying levels of temporal dependency structure. SW represents the amount of water in the soil that changes due to evaporation, plant intake, etc. Among these three variables, SW has the longest temporal dependency structure spanning multiple years. SNO represents the amount of water available in the form of snow. SNO temporal dependency is seasonal because snow completely melts by the end of summer in many locations. In other words, unlike SW, SNO state resets to zero every year. Finally, SF is a flux variable whose temporal dependency can largely be removed due to major weather events like precipitation and snowmelt. In addition, SF also depends on the value of state variables and weather drivers. For example, wet soil will lead to more SF for the same amount of rainfall or snowmelt compared to dry soil. Hence, these three variables provide a good framework for analyzing different training strategies. Figure \ref{fig:inputs_7feats_outputs_3feats} shows the timeseries of weather inputs and these three target variables. This paper shows results using a SWAT simulation for a watershed in Southwest Minnesota. We provide results on additional six watersheds in the appendix.

\subsection{Experimental Setup.} \label{sec:experiments_setup}
The time series data ranges from 1902-01-01 to 2902-01-01 with a total of 365244 days. We use the first 182622 days ($50\%$ series) as the training set, the middle 36524 days ($10\%$ series) as the validation set, and the last 146098 days ($40\%$ series) as the testing set. We apply gaussian normalization on the training set, then use the calculated sample means and variances to normalize the validation and testing sets. We compare GRU, LSTM, and Transformer (section \ref{sec:base_models}) and decide to use GRU as the base model to make many-to-many predictions.  we slice the normalized training, validation, and testing time series into segments of length 366. Every consecutive segment does overlap with 183 days in RMB-IIF and CMB-SCIF strategies and does not overlap in SMB-SSIF and SSMB-SSIF strategies. 

The learning methods train models on the mean square error (MSE) loss function using the Adam algorithm \cite{kingma2014adam} in the default PyTorch setting except for the $0.01$ learning rate. In each epoch of RMB and CMB, we shuffle the segments in the training set and assign 64 segments in each batch to train the models. In SMB and SSMB algorithms, we organize segments in specific orders to allow hidden states to pass between batches or segments. We train models 500 epochs in maximum and stop the training if the models can not get a smaller MSE within consecutive 50 epochs, called early stopping. Finally, we use the pre-calculated mean and variance of the target variable to transform the predictions back to the original scale before evaluation. We examine the learning strategies 5 times to train models with different initial weights to quantify the stability of the strategies. 

% %%%%%% newly added subsection
\subsection{Performance metrics}
\subsubsection{Mean square error (MSE).} \label{sec:MSE}
We use MSE as the loss function to train models. MSE is the sum of the squares of residuals to measure the discrepancy between the predictions and observations, widely used in ML and statistics. MSE ranges from 0 to $+\infty$; a smaller MSE indicates the model fits the data better.  

\subsubsection{Root mean square error (RMSE).}
RMSE is simply the square root of the MSE. The square root keeps RMSE on the same scale as the observations, making RMSE more interpretable for the domain scientists. RMSE ranges from 0 to $+\infty$, and the smaller RMSE means better predictions. 
\begin{align}
RMSE(Y, \hat{Y}) = \sqrt{\frac{1}{N}\sum_{i=1}^{N}(\hat{Y}^i - Y^i)^2} \label{eq:RMSE}
\end{align}

\subsubsection{Nash–Sutcliffe model efficiency coefficient (NSE).}
NSE is widely used to measure the predictive skill of hydrologic models. Equation (\ref{eq:NSE}) shows the NSE equals one minus the ratio of error variance (MSE) divided by the variance of the observations. When the error variance is larger than the variance of the observations, NSE becomes negative. As a result, NSE ranges from $-\infty$ to 1. The higher NSE indicates better prediction skills of the models. 
\begin{align}
NSE(Y, \hat{Y}) = 1 - \frac{\sum_{i=1}^{N}(\hat{Y}^i - Y^i)^2}{\sum_{i=1}^{N}(Y^i - \bar{Y})^2} \label{eq:NSE}
\end{align}
% %%%%%%

\section{Results}
Table \ref{tab:rep7_GRU_vs_SPGRU_SW} summarizes the RMSE and NSE values of predictions on the testing set for all three variables. We can see the trends in performance vary depending on the temporal dependency on variables. For SW and SNO, both proposed strategies outperform existing baselines. However, CMB-SCIF performs similarly to RMB-IIF for SF. Since SF is a flux variable whose values and temporal dependencies can be largely reset by weather inputs like precipitation and snowmelt, access to the initial value does not provide additional information. SSMB-SSIF can slightly improve SF because SSMB-SSIF uses hidden states which contain information in addition to the target values that help SF modeling. The table also shows the impact of sharing information between segments during inference. Specifically, we observe that SSIF significantly improves performance compared to IIF when they both use the same RMB-trained models in inference.

\begin{table*}[h]
    \caption{The table shows the overall performance of models from different learning methods.}
    \label{tab:rep7_GRU_vs_SPGRU_SW}
    \resizebox{\textwidth}{!}{% Resize table to fit within \linewidth
    \begin{tabular}{ |c|c|c|c|c|c|c|}
        \hline
        &\multicolumn{2}{|c|}{SW} & \multicolumn{2}{|c|}{SNO} & \multicolumn{2}{|c|}{SF}\\
        \hline
        Strategy & RMSE& NSE & RMSE& NSE& RMSE& NSE\\
        \hline
        RMB-IIF  & $32.796 \pm 0.229$ &$0.627 \pm 0.005$ &$2.588 \pm 0.06$ & $0.952 \pm 0.002$ &$0.663 \pm 0.013$ & $0.948 \pm 0.002$\\
        RMB-SSIF & $18.92 \pm 0.857$ &$ 0.876 \pm 0.011$ &$1.645 \pm 0.153$ & $0.98 \pm 0.004$&$0.522 \pm 0.044$ & $0.968 \pm 0.006$\\
        SMB-SSIF & $17.875 \pm 1.203$ &$0.889 \pm 0.015$ &$1.556 \pm 0.281$ & $0.982 \pm 0.007$ & $0.525 \pm 0.033$ & $0.968 \pm 0.004$\\
        % SMB-SSIF(bs1) & $24.005 \pm 1.41$ &$0.799 \pm 0.024$ &$3.844 \pm 0.434$ & $0.893 \pm 0.024$ & $0.769 \pm 0.064$ & $0.93 \pm 0.012$\\
        SSMB-SSIF & $\bm{15.055 \pm 1.282}$ &$\bm{0.921 \pm 0.013}$ &$1.364 \pm 0.059$ & $0.987 \pm 0.001$&$\bm{0.492 \pm 0.029}$ & $\bm{0.971 \pm 0.003}$\\
        CMB-SCIF & $16.272 \pm 0.367$ &$0.908 \pm 0.004$ &$\bm{1.099 \pm 0.104}$ &$\bm{0.991 \pm 0.002}$&$0.678 \pm 0.02$ &$0.946 \pm 0.003$ \\
        \hline
    \end{tabular}
    }
\end{table*}

\subsection{Average RMSE at each timestep of RNN segments.} 
\begin{figure}[tb]
  \centering
  \includegraphics[width=0.35\columnwidth]{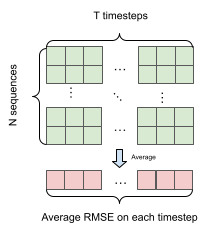}
  \includegraphics[width=0.35\columnwidth]{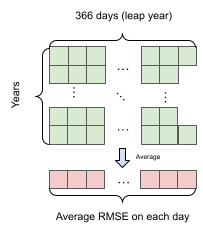}
  \caption{The left plot explains how we compute the average RMSE at each timestep of RNN segments. The right plot explains how we compute average RMSE at each day across years.}
  \label{fig:timestep_daily_rmses_explain}
\end{figure}

\begin{figure}[h]
  \centering
  \includegraphics[width=\columnwidth]{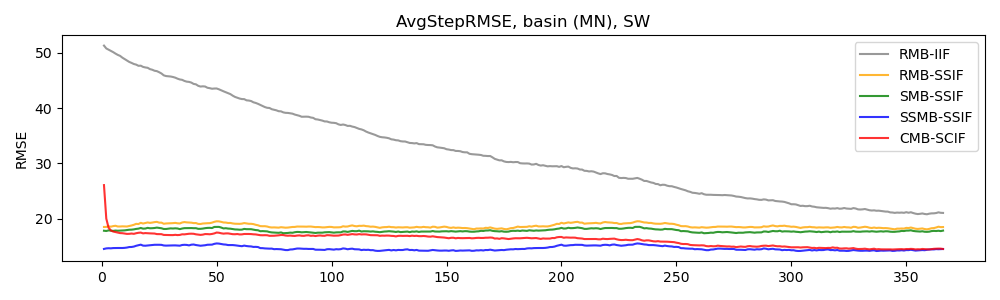}
  \includegraphics[width=\columnwidth]{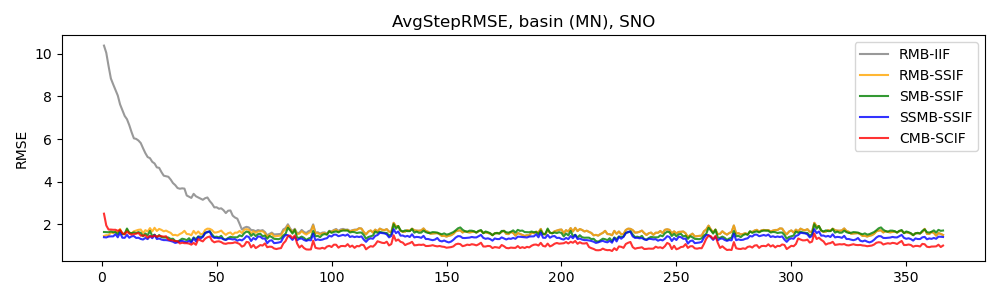}
  \includegraphics[width=\columnwidth]{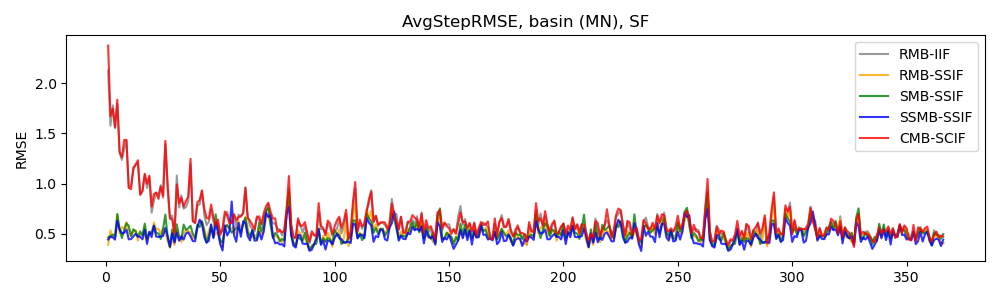}
  \caption{This figure shows the average RMSE at each RNN timestep, explained in fig \ref{fig:timestep_daily_rmses_explain}, on testing set.}
  \label{fig:rep7_timestep_rmses}
\end{figure}

In this study, the learning strategies train GRUs to make predictions at every timestep. As the left plot of figure \ref{fig:timestep_daily_rmses_explain} explains, we compute the average RMSE at each timestep (AvgStepRMSE) on the testing set to compare the learning strategies at different timesteps within the segments. We can use AvgStepRMSE to measure how many timesteps different strategies require before they can generate reasonable predictions.

As figure \ref{fig:rep7_timestep_rmses} shows, RMB-IIF learning gets a very high AvgStepRMSE at the early steps, then slowly decreases AvgStepRMSE, which reveals that RMB-IIF learning needs longer timesteps to build informative hidden states. On the contrary, RMB-SSIF, SMB-SSIF, and SSMB-SSIF get flat AvgStepRMSE traces because they do not reset the hidden states between segments. The same reason explains why RMB-SSIF, which uses SSIF inference, can perform much better than RMB-IIF learning though they use the same trained model. 

CMB-SCIF learning has a distinct AvgStepRMSE pattern. In the SF task, CMB-SCIF performs no differently than RMB-IIF. In SW and SNO tasks, CMB-SCIF gets intermediate AvgStepRMSE at early time steps, quickly decreases AvgStepRMSE within a few timesteps, and keeps reducing the AvgStepRMSE to even lower values until the last timestep. For both SW and SNO tasks, proposed mini-batch strategies improve performance even on the last timestep compared to RMB. This highlights the utility of incorporating temporal dependency to improve performance on all timesteps and not just earlier timesteps. 

\subsection{Average RMSE at each day across years.}
\begin{figure}[tb]
  \centering
  \includegraphics[width=\columnwidth]{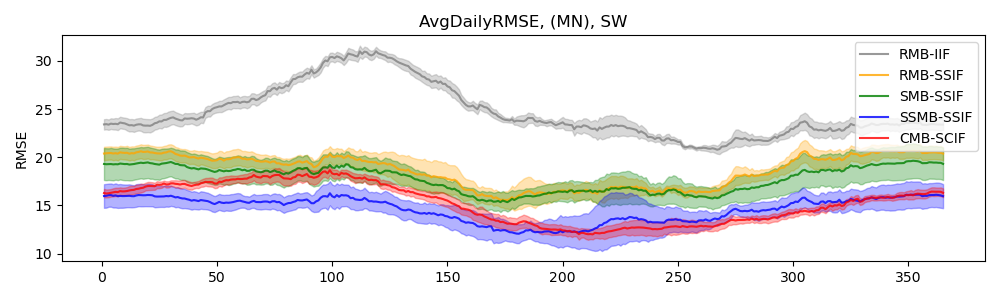}
  \includegraphics[width=\columnwidth]{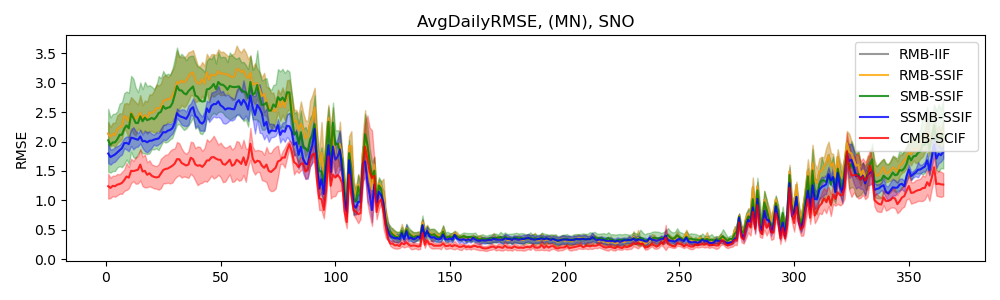}
  \includegraphics[width=\columnwidth]{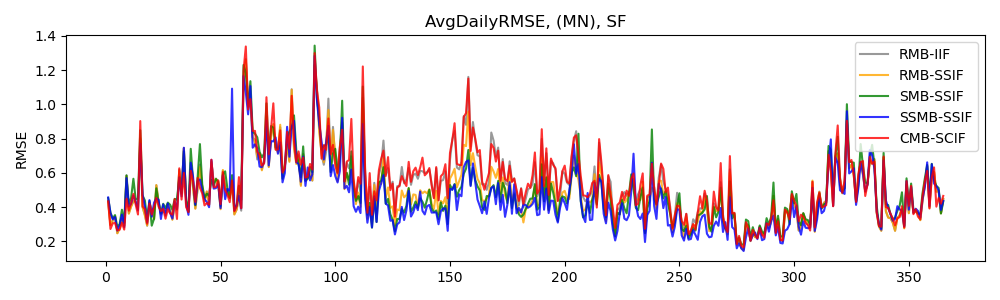}
  \caption{This figure shows the traces of average daily RMSE across years, explained in fig \ref{fig:timestep_daily_rmses_explain}, on the testing set. The shaded regions, muted in the SF plot for clarity, cover one standard deviation from five runs on models with different initial weights.}
  \label{fig:rep7_daily_rmses}
\end{figure}

We analyze the average RMSE each day across years (AvgDailyRMSE), as right plot of figure \ref{fig:timestep_daily_rmses_explain} explains. 
We can use AvgDailyRMSE to reveal seasonal patterns in the loss of trained models on SW, SNO, and SF tasks. As shown in Figure \ref{fig:rep7_daily_rmses}, our proposed learning strategies perform better than baselines on all days of the year for the SW and SNO tasks. For SNO, AvgDailyRMSE traces of the baselines, RMB-IIF, RMB-SSIF, and SMB-SSIF's AvgDailyRMSE, are primarily overlapped since RMB-IIF provides as good SNO predictions as other baselines after 100 timesteps as shown in Figure \ref{fig:rep7_timestep_rmses}. 

The simulated watershed is a snow-dominated watershed that exhibits a strong seasonal pattern of snow accumulation. The SNO variable is zero in summer but is more active and dynamic in other seasons when either snow falls or melts. This seasonal pattern can be observed in our prediction results in Figure \ref{fig:rep7_daily_rmses}. 
The AvgDailyRMSE traces of SNO get lower values in summer since snow only happens when the temperature is below the freezing point. 

Snow is a type of precipitation that has delayed effects on SW and SF. Figure \ref{fig:rep7_daily_rmses} shows higher AvgDailyRMSEs for both SW and SF around days 50 to 150 when the snow starts melting in spring. Except RMB-IIF learning, all other learning strategies get relatively low AvgDailyRMSE in spring on SW prediction, showing that passing information between segments helps models capture snow effects better. 

On SF prediction, although all strategies get similar AvgDailyRMSE in winter and spring, RMB-SSIF, SMB-SSIF, and SSMB-SSIF get lower AvgDailyRMSE in summer, showing that hidden states could carry helpful information about SW and capture the effect of snow melting on SF prediction in summer. 

\subsection{Time efficiency.}

\begin{table}[ht]
    \caption{This table records how many seconds the learning methods need to train a model in an epoch.}
    \label{tab:training_epoch_time}
    \resizebox{\columnwidth}{!}{% Resize table to fit within \linewidth
    \begin{tabular}{|c|c|c|c|c|c|}
        \hline
         & RMB & SMB& SSMB & CMB \\
        \hline
        sec/epoch & 0.0172& 0.0211& 0.5390 & 0.0151\\
        \hline
    \end{tabular}
    }
\end{table}
Table \ref{tab:training_epoch_time} shows that SSMB takes 0.5390 seconds to finish an epoch on an NVIDIA A100 GPU since SSMB has to process segments one by one and wait for hidden states from the previous to initialize the current segment. RMB, SMB, and CMB algorithms only use around 0.017 seconds since they process segments within a batch independently while leveraging the GRUs' matrix acceleration to reach higher time efficiency. This shows that CMB can achieve good performance improvements with minimal effect on training times compared to SSMB.

\subsection{Data efficiency.}
\begin{figure}[h]
  \centering
  \includegraphics[width=\columnwidth]{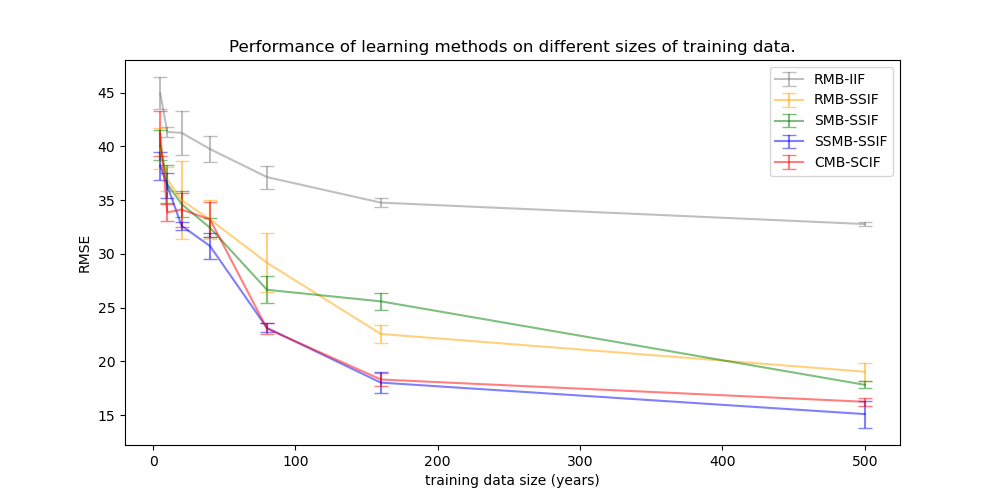}
  \caption{We vary the training data sizes (5, 10, 20, 40, 80, 160, and 500 years) and use different learning methods to train models on SW modeling. We use the trained models to predict the same testing set and get RMSE traces in this figure.}
  \label{fig:rmses_training_size_sw}
\end{figure}

To evaluate the methods' generalizability, we assess how much training data the learning methods require to train models on SW prediction. As Figure \ref{fig:rmses_training_size_sw} shows, the relative position of the traces is consistent with the previous results where the proposed SSMB-SSIF and CMB-SCIF strategies perform better than baseline across all training data sizes. The SSMB-SSIF and CMB-SCIF get close RMSE traces, showing that both initial target values and hidden states can help simulate SW.

\subsection{Impact of inference initialization on SCIF.}
\begin{figure}[h]
  \centering
  \includegraphics[width=\columnwidth]{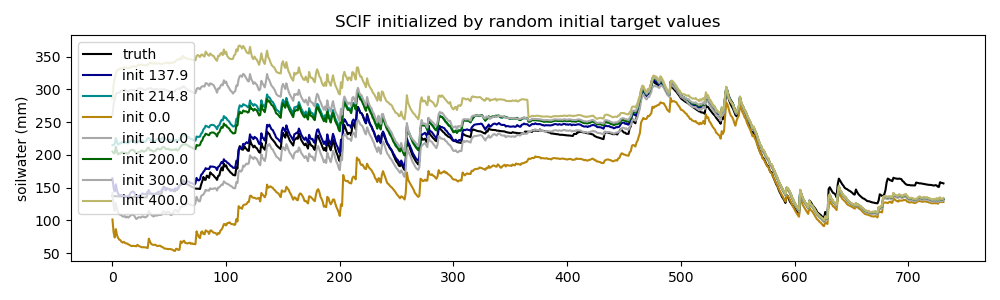}
  \includegraphics[width=\columnwidth]{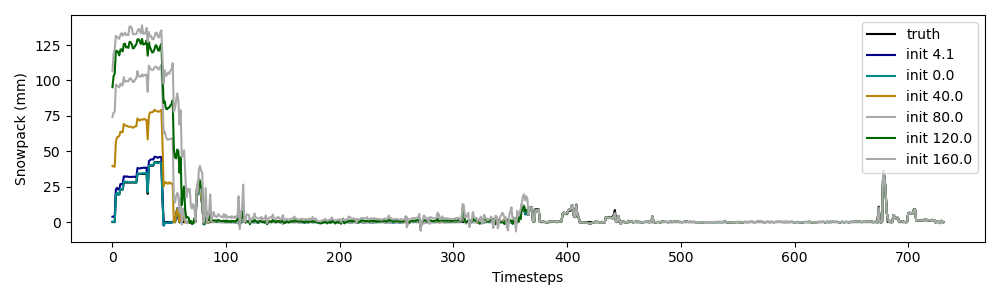}
  \caption{This figure shows the predicted segments from SCIF inference algorithm initialized by different initial target values.}
  \label{fig:SP_random_inits}
\end{figure}

Figure \ref{fig:SP_random_inits} shows the predicted SW and SNO time series from SCIF inference initialized by varying initial targets. We provide 7 SW initial values, including the observed value (137.9),  the mean values (214.8) from training data, and five extrapolated values (0, 100, 200, 300, 400) from the range of SW. Figure \ref{fig:SP_random_inits} shows that all seven SW predictions keep merging and finally overlapping each other within 600 time steps. We initialize the SCIF inference using 6 SNO values, including the mean values (4.1) from training data and the five extrapolated values (0, 40, 80, 129, 160) from the range of SNO (0 happens to be the observed initial value). Figure \ref{fig:SP_random_inits} shows all six SNO predictions merge within 100 time steps.

These predicted SNO time series merge faster because SNO resets to 0 in summer. These results prove that SCIF inference is sensitive to initial memory status during early-stage simulations but will ultimately converge to reflect the dominant impacts from weather drivers.

\subsection{Comparisons with teacher forcing.}
\begin{figure}[h]
  \centering
  \includegraphics[width=\columnwidth]{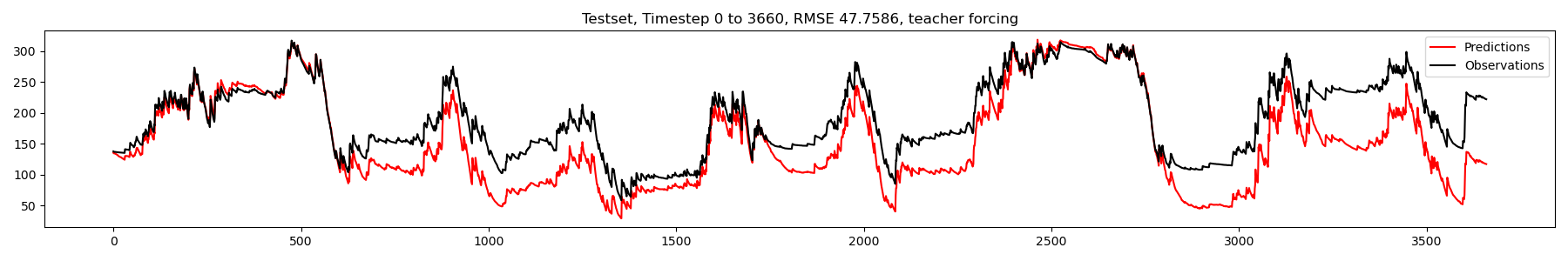}
  \includegraphics[width=\columnwidth]{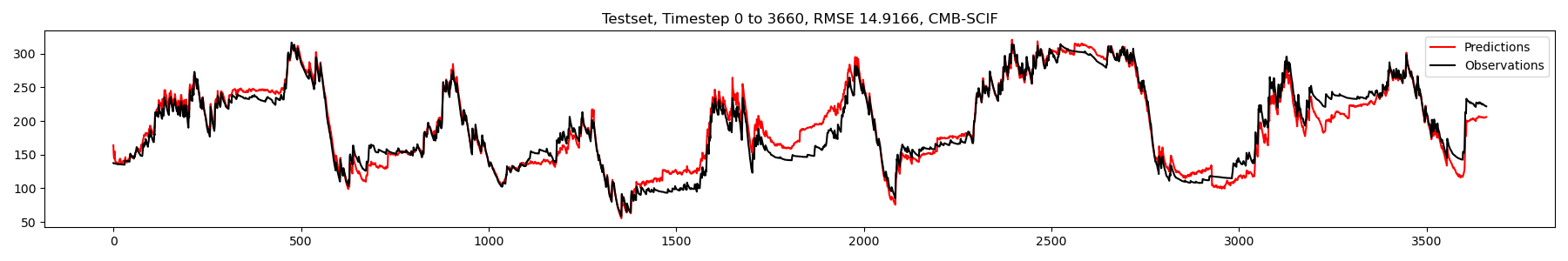}
  \caption{Predicted timeseries of teacher forcing and CMB-SCIF on SW task.}
  \label{fig:SP_teacher_forcing}
\end{figure}

As discussed earlier, CMB-SCIF is a special case of teacher forcing strategy. Table \ref{tab:teacher_forcing} shows CMB-SCIF performs consistently better than teacher forcing on SW, SNO, and SF tasks. Since SNO completely melts by the end of summer and  SF is strongly influenced by precipitations, the error propagation can be largely reset and be under controlled; hence, teacher forcing performs only slightly worse than CMB-SCIF. However, teacher forcing, although having to use an observed initial value to start inference, performs much worse on SW since the error propagation issue is out of control due to the long-term temporal dependency of SW, as Figure \ref{fig:SP_teacher_forcing} reveals.

\begin{table}[h]
    \caption{The table compares the testing RMSEs.}
    \label{tab:teacher_forcing}
    \resizebox{\columnwidth}{!}{% Resize table to fit within \linewidth
    \begin{tabular}{ |c|c|c|c|}
        \hline
        Strategy & SW & SNO & SF\\
        \hline
        % RMB-IIF  & $32.796 \pm 0.229$ & $2.588 \pm 0.06$ &$\bm{0.663 \pm 0.013}$ \\
        Teacher forcing & $34.31 \pm 11.27$ &$1.23 \pm 0.52$ &$0.8 \pm 0.29$\\
        CMB-SCIF & $\bm{16.272 \pm 0.367}$ &$\bm{1.099 \pm 0.104}$ &$\bm{0.678 \pm 0.02}$\\
        \hline
    \end{tabular}
    }
\end{table}

\section{Conclusion} \label{sec:conclusion}
In this paper, we proposed to use hidden states and initial target values to incorporate temporal dependence between training segments for timeseries prediction tasks. Specifically, we introduce RMB-SSIF learning to allow RMB-trained models to pass hidden states in the testing phase, preventing models from building up informative hidden states from scratch for each segment. We further show that RMB-SSIF learning performs similarly to SMB-SSIF learning. To deal with the weakness of SMB-SSIF learning, we design the SSMB-SSIF learning to train models using all historical information while maintaining more segments in each batch to stabilize the optimization. We show that SSMB-SSIF learning performs consistently better than SMB-SSIF learning on SW, SNO, and SF predictions. We further show that CMB-SCIF can provide similar improvements while maintaining faster training times similar to traditional mini-batch learning. For future work, we aim to test these strategies on other timeseries applications such as finance and healthcare. 

\section{Acknowledgments}
The work is being funded by NSF HDR grant 1934721, 1934548, NSF FAI grant 2147195, NASA award 80NSSC22K1164, the USGS awards G21AC10207, G21AC10564, and G22AC00266. Access to computing facilities was provided by Minnesota Supercomputing Institute.

\bibliographystyle{unsrt}
\bibliography{references.bib}

\appendix
\section{Base models selection}\label{sec:base_models}
Table \ref{tab:base_models} shows that GRU performs similarly to the LSTM and better than the Transformer. Since GRU has simpler architecture, we use GRU as the base model to measure the performance of the learning strategies.

\begin{table}[ht]
    \centering
    \caption{The settings of base models.}
    \label{tab:base_models_settings}
    \resizebox{\columnwidth}{!}{% Resize table to fit within \linewidth
    \begin{tabular}{|c|c|}
    \hline
    Model & Settings\\
    \hline
    GRU &1-layer, 32 hidden state size\\
    \hline
    LSTM &1-layer, 32 hidden state size\\
    \hline
    Transformer &\makecell{1-layer, 21 embedding size,7 head numbers,\\ masked multi-head attention}\\
    \hline
    \end{tabular}
    }
\end{table}
\begin{table}[ht]
    \caption{The overall RMSEs of base models on RMB-IIF strategy.}
    \label{tab:base_models}
    \resizebox{\columnwidth}{!}{% Resize table to fit within \linewidth
    \begin{tabular}{ |c|c|c|c|}
        \hline
        &SW &SNO &SF \\
        \hline
        GRU  & $32.796 \pm 0.229$ &$2.588 \pm 0.06$ &$0.663 \pm 0.013$\\
        \hline
        LSTM  & $33.487 \pm 0.406$ &$2.483\pm 0.067$ &$0.685\pm 0.011$\\
        \hline
        Transformer & $35.400 \pm 0.174$ & $6.818 \pm 0.555$ &$1.301 \pm 0.014$\\
        \hline
    \end{tabular}
    }
\end{table}

\begin{figure}[h]
    \centering
    \includegraphics[width=0.9\columnwidth]{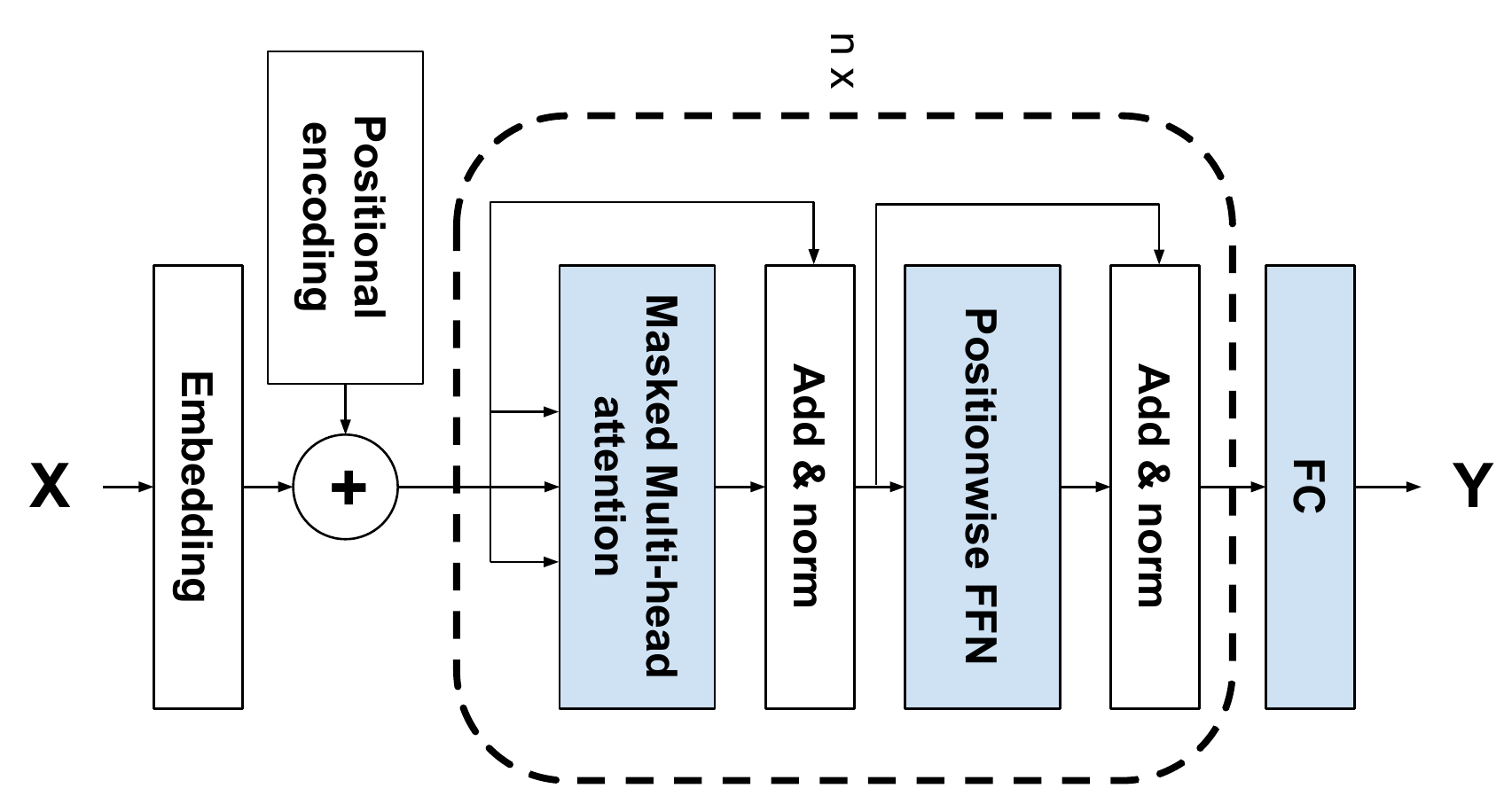}
    \caption{The investigated transformer model \cite{https://doi.org/10.48550/arxiv.1706.03762,kratzert2022joss}.}
    \label{fig:maksed_encoder}
\end{figure}

\section{The additional six watersheds}
\begin{table}[h!]
  \caption{basin characteristics}
  \label{tab:basins_characteristics}
  \resizebox{\linewidth}{!}{% Resize table to fit within \linewidth horizontally
\begin{tabular}{ |c|c|c|c|c|}
 \hline
 Basin & state & mean precipitation  &area gauged & frac\_snow\\
    &&$mm$&$km^2$&$\%$\\
 \hline
    % MN &MN& $2.300419$ &$11.12$&$0.298$\\
    01162500 &MA& $3.783534565$ &$49.71$&$0.226769185$\\
    06614800 &CO& $2.917613963$ &$4.03 $&$0.760476738$\\
    12114500 &WA&$7.241880903$& $66.56$ &$0.406517039$\\
    02300700 &FL&$3.89649692$&$73.84$ & $0$\\
    11141280 &CA&$1.883665982$ &$54.01$ & $0.00078778$\\
    09066200 &CO&$2.650229979$ &$16.1$ & $0.712216418$\\
 \hline
\end{tabular}
  }
\end{table}

We provide results on six additional basins which span drought, snow, and humid regions across the USA, as table \ref{tab:basins_characteristics} shows.

\begin{figure*}[t]
\centering
\includegraphics[width=0.97\textwidth]{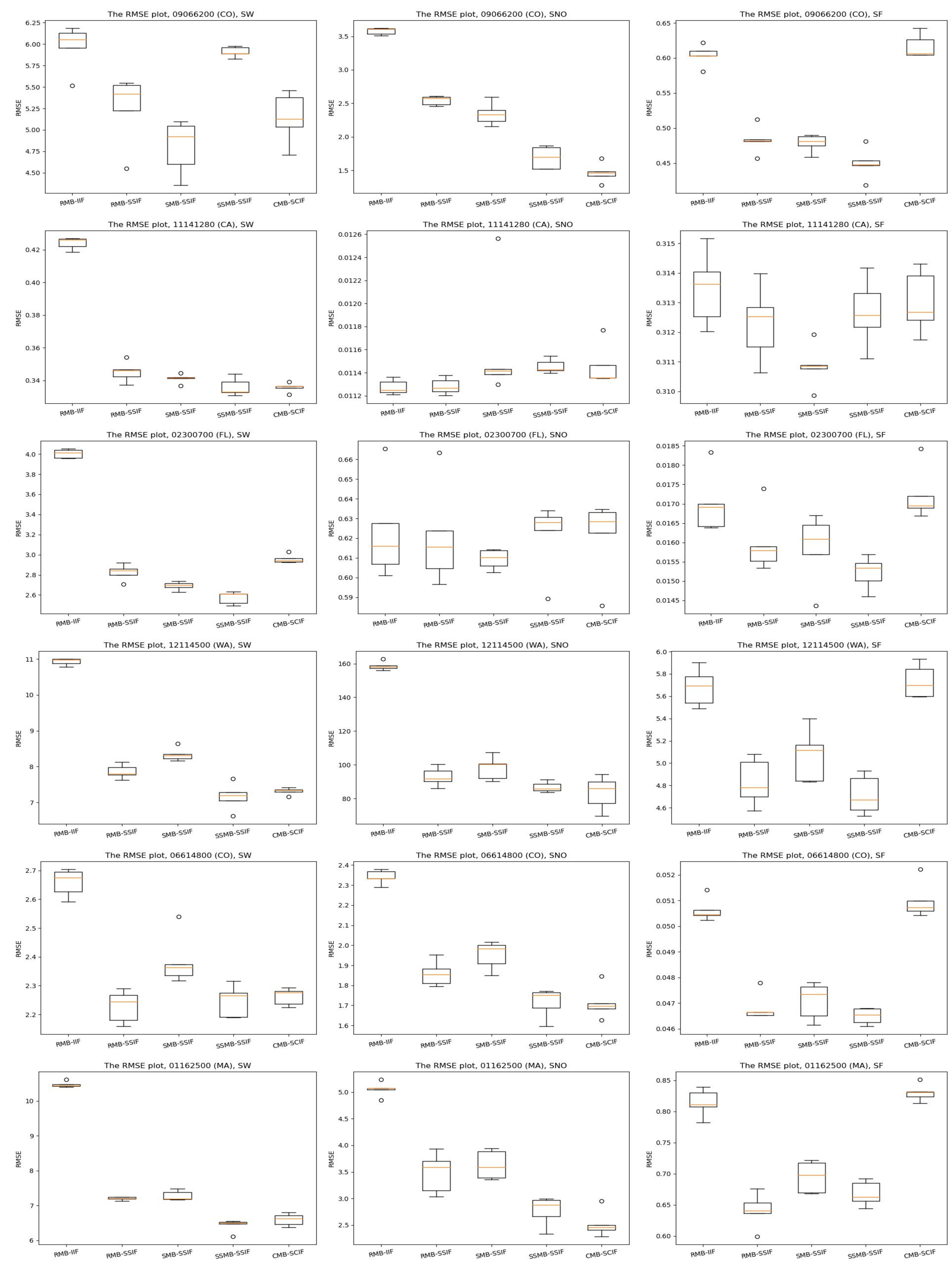}
\caption{These boxplots draw the overall RMSEs of learning methods on different basins.}
\label{fig:boxplots_rmse}
\end{figure*}

\begin{figure*}[t]
\centering
\includegraphics[width=\textwidth]{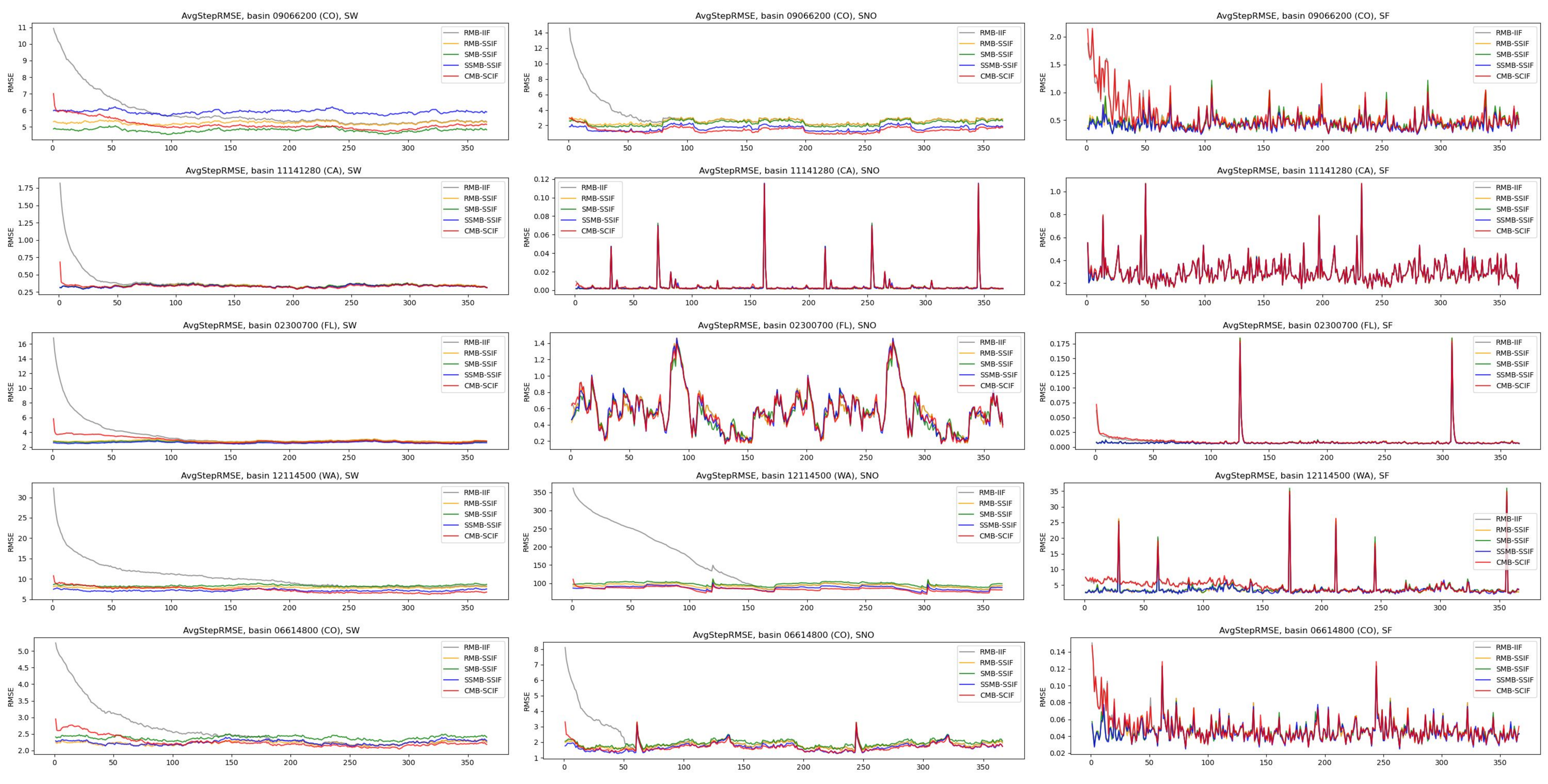}
\caption{These figure shows the average RMSE at each RNN timestep on testing sets across all different basins.}
\label{fig:avg_step_rmse_basins_set1}
\end{figure*}

Figure \ref{fig:boxplots_rmse} shows the boxplots of the overall RMSEs from 5 separate runs of the learning strategies on the basins. And Figure \ref{fig:avg_step_rmse_basins_set1} are the AvgStepRMSE plots. All figures show that the behaviors of the learning methods are mostly consistent across different basins and conform to the findings in the paper but with few exceptions. 

\begin{figure*}[t]
\centering
\includegraphics[width=0.32\textwidth]{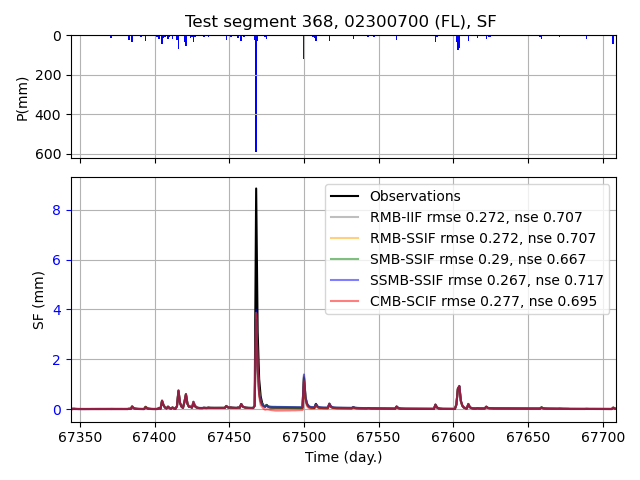}
\includegraphics[width=0.32\textwidth]{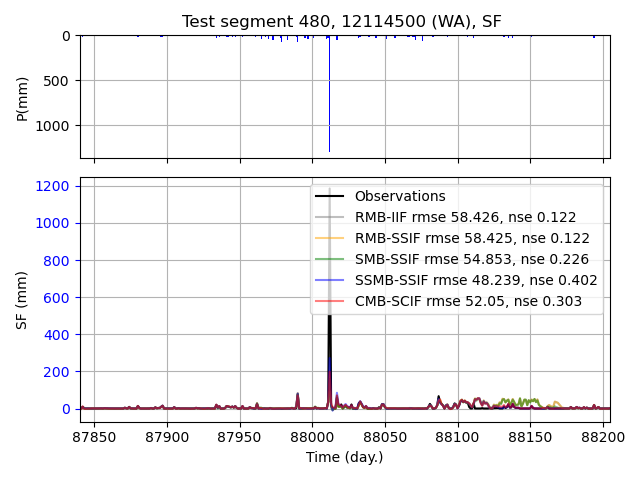}
\includegraphics[width=0.32\textwidth]{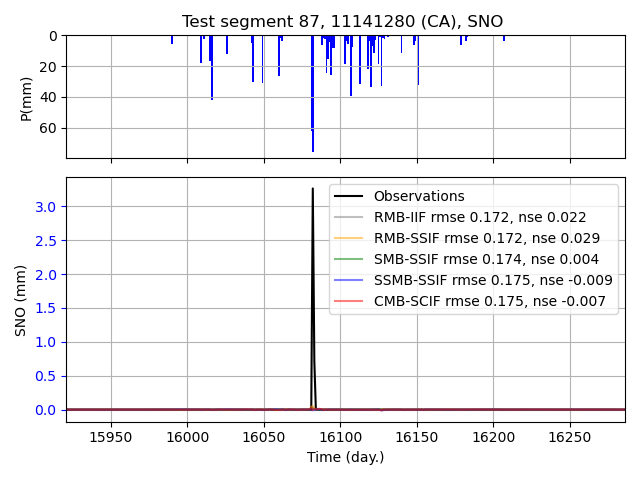}
\caption{This figure shows three hydrographs. The top plot of the hydrograph draws the precipitation (P); the bottom plot draws the target variables.}
\label{fig:trouble_shoot_hydrograph}
\end{figure*}

The figure \ref{fig:avg_step_rmse_basins_set1} shows all strategies get some abnormal AvgStepRMSE peaks in SF predictions on basin 02300700 at FL and basin 12114500 at WA and in SNO predictions on basin 11141280 at CA. These abnormal peaks happen due to some extremely rare floods and snowstorms. We draw the hydrograph to study these extreme events in figure \ref{fig:trouble_shoot_hydrograph}, which reveal the extreme rainstorm and snowstorm make the observations several magnitudes higher than usual. Since such events only happen once in several decades, all strategies underpredict values, so we observe the AvgStepRMSE peaks. 

\end{document}